\documentclass{article}

\usepackage{PRIMEarxiv}

\usepackage[utf8]{inputenc} 
\usepackage[T1]{fontenc}    
\usepackage{url}            
\usepackage{booktabs}       
\usepackage{amsfonts}       
\usepackage{nicefrac}       
\usepackage{microtype}      
\usepackage{fancyhdr}       
\usepackage{graphicx}       
\usepackage{float}
\graphicspath{{media/}}     
\usepackage{hyperref}       
\usepackage{multirow}

\usepackage{amsmath}

\title{Discriminative Span as a Predictor of Synthetic Data Utility via Classifier Reconstruction}

\author{
  Radhika Amar Desai\\
  School Of Computer Science \\
  Vellore Institute of Technology \\
  Chennai\\
  \texttt{radhikaamar.desai2022@vitstudent.ac.in} \\
   \And
  Modigari Narendra \\
  School Of Computer Science \\
  Vellore Institute Of Technology \\
  Chennai\\
  \texttt{modigari.narendra@vit.ac.in} \\
}

\begin{document}
\maketitle

\begin{abstract}

In many real-world computer vision applications, binary classification tasks are characterized by a severe scarcity of positive samples. A widely adopted solution is to generate synthetic positive data using image to image transformations applied to negative samples. However, how can we reliably assess whether such synthetic data will improve downstream model performance is a fundamental challenge. In this work, we discuss a geometry driven metric that predicts the utility of synthetic data without requiring model training. Our approach operates in the embedding space of a pre-trained foundation model and represents the dataset through difference vectors between samples. We evaluate whether the weight vector of a linear classifier can be expressed within the subspace spanned by these variations measuring the relative projection error. Intuitively, if the variations induced by synthetic data capture tasks relevant directions, their span can approximate the classifier, resulting in a low projection error. Conversely, poor synthetic data fails to span these directions, leading to higher error. Using five datasets and three architectures, we show that this metric exhibits strong correlation with the downstream classification performance of neural networks trained on mixtures of real negative and synthetic positive data. These findings suggest that the proposed metric serves as a practical and informative tool for evaluating synthetic data quality in data scarce settings.

\end{abstract}

\section{Introduction}

In computer vision tasks like medical imaging or industrial inspection, we constantly run into the same roadblock: a severe shortage of positive training samples. This imbalance naturally hurts model performance and reliability. A common workaround is to use Image to Image (I2I) translation to generate synthetic data, boosting diversity across domains~\cite{zhu2017unpaired,wolterink2017deep}. Yet, a critical question remains: how do we actually know if this synthetic data will help the downstream model? Right now, the standard way to evaluate synthetic data is incredibly tedious: you train a model on the augmented dataset and see how it performs on a test set~\cite{figueira2022survey}. This approach is not only computationally expensive and highly sensitive to training quirks, but also fails to explain why certain synthetic data works while others fail. A more principled, lightweight way to predict synthetic data utility without burning compute on full model training is needed. This work tackles this from a geometric angle, looking at whether the variations introduced by synthetic data actually capture task relevant directions in the representation space. The core idea is to see if we can reconstruct the decision boundary of a target task using just the span of these data variations. If the synthetic data adds meaningful and task aligned variations, those variations should align closely with the classifier's direction. To do this, we work inside the embedding space of a pre-trained foundation model and represent the dataset using difference vectors between samples.We then measure the relative projection error to assess how well the weight vector of a linear classifier aligns with the subspace spanned by these variations, a quantity we call Discriminative Span (DS). This approach is validated across five datasets and three architectures, finding a strong empirical correlation between DS and the actual downstream performance of CNNs trained on real negative samples and their corresponding synthetic positive data samples. The DS offers a practical, predictive proxy for vetting synthetic data quality when real data is hard to come by.

Our main contributions are as follows:
\begin{itemize}
    \item Introduces \textit{Discriminative Span}, a geometry-driven metric for predicting the utility of synthetic data without requiring model training.
    \item Proposes a formulation based on reconstructing the classifier direction from the span of data-induced variations in representation space.
    \item Demonstrates strong empirical correlation between the proposed metric and downstream classification performance across multiple datasets and architectures.
    \item Provides ablation studies and analysis to validate the design choices and robustness of the proposed metric.
\end{itemize}

\section{Related Work}
Synthetic data generation has become a common strategy for addressing data scarcity in computer vision, particularly through I2I translation. Unpaired methods such as CycleGAN~\cite{zhu2017unpaired} enable generation without paired training data, and subsequent approaches have improved fidelity and controllability through architectural and training modifications~\cite{wang2022dccyclegan}. Existing evaluation methods primarily assess visual fidelity, distributional similarity or downstream utility using common metrics like SSIM, PSNR, Fréchet Inception Distance (FD) and Inception Score, while task specific evaluation typically requires training a downstream model on the generated data~\cite{figueira2022survey,abdusalomov2023evaluating,mumuni2024survey}. Recent surveys and evaluation frameworks emphasize that fidelity and distributional similarity do not necessarily imply downstream utility, particularly for specialized domains such as medical imaging~\cite{dankar2022multidimensional,ibrahim2025generative, koetzier2024synthetic,zamzmi2025scorecard, sizikova2024synthetic}. This work predicts downstream utility directly from the geometry of synthetic data induced variations, without training a downstream model.

Difference vectors in representation spaces have been widely used to characterize and manipulate semantic attributes. Vector arithmetic in learned latent spaces can capture interpretable transformations, while disentangled I2I approaches use latent directions to control attributes and generate diverse outputs~\cite{radford2015unsupervised, lee2018diverse}. Methods such as VecGAN and SliderGAN further demonstrate that meaningful visual transformations can be represented through directions in learned latent spaces~\cite{vecgan2022, slidergan2020}. These works motivate our use of real-to-synthetic embedding differences as representations of transformation-induced variation. Rather than using such directions for generation or editing, we investigate whether their span contains the task-relevant discriminative structure.

Linear probing is widely used to evaluate whether learned representations encode linearly separable task information~\cite{chen2020simple}. More broadly, representation-learning and transfer-learning methods rely on the geometry of pretrained embedding spaces to analyze class structure and transferability. Our approach builds on this geometric perspective but considers a different object: instead of measuring whether synthetic and real samples are close, or whether classes are linearly separable, we measure whether the span of synthetic-induced difference vectors can reconstruct the discriminative direction of the target task. This distinction allows Discriminative Span to explicitly evaluate task-relevant variation rather than global fidelity or static separability.

\section{Method}
\begin{figure}[t]
    \centering
    \includegraphics[width=0.9\linewidth]{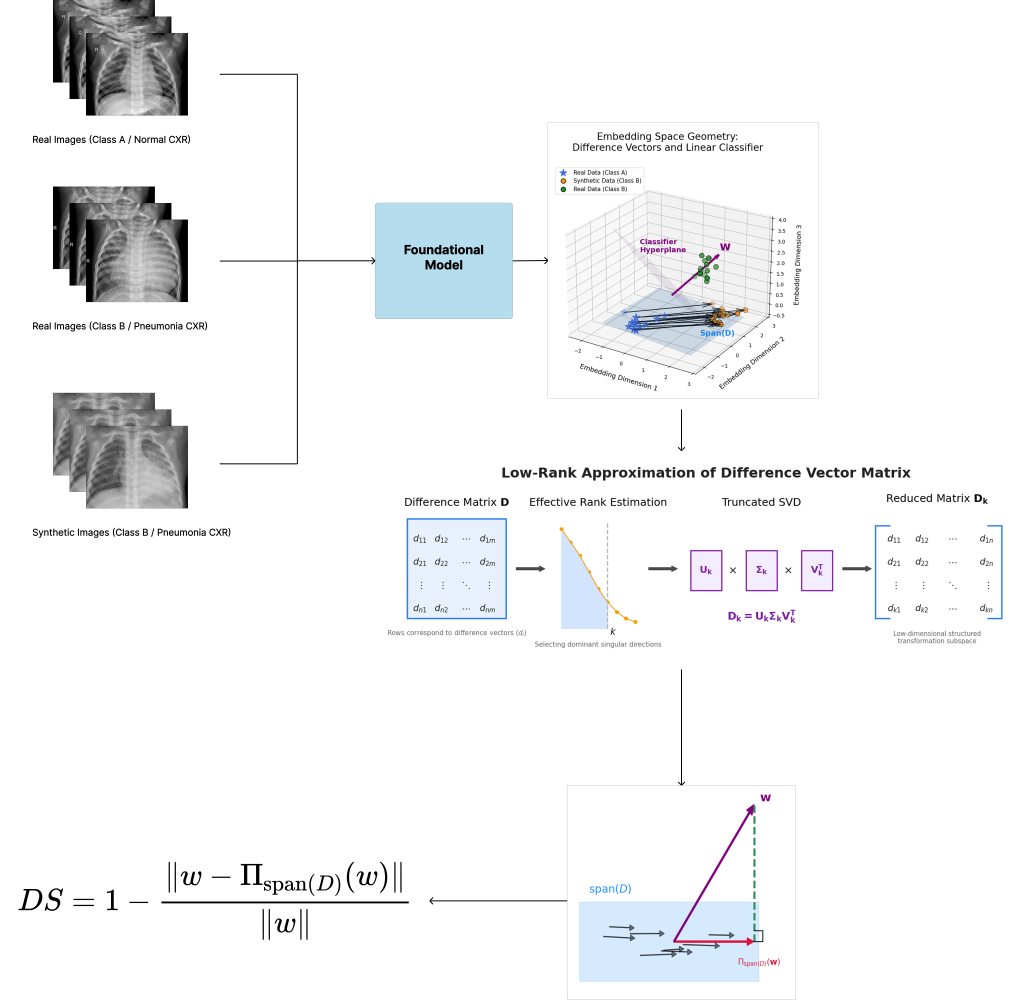}
    \caption{ Overview of the proposed framework for analyzing synthetic data’s model utility in embedding space.}
    \label{fig:overall_pipeline}
\end{figure}

Figure \ref{fig:overall_pipeline} presents the framework proposed for analyzing synthetic data's model utility in embedding space. The images are first mapped into latent representations using a foundational vision model. Difference vectors constructed between embeddings are organized into a difference matrix D, whose dominant structure is extracted through effective rank estimation and truncated SVD to obtain a low-rank approximation. A linear classifier defines a discriminative direction $w$ in the embedding space, and the proposed Discriminative Span (DS) metric measures how well $w$ aligns with the span of the reduced difference vector space. Higher DS values indicate better preservation of task relevant discriminative structure by the synthetic data.

\subsection{Problem Setup}
We consider a binary classification task in the embedding space of a pre-trained foundation model. Let $x \in \mathbb{R}^d$ denote an image embedding and let a linear classifier be defined by
\[
w^T x + b = 0,
\]

where $w \in \mathbb{R}^d$ is the classifier weight vector. Since $w$ is normal to the decision boundary, it is interpreted as the discriminative direction that captures the task relevant signal separating the two classes. The classifier is trained using embeddings of real positive and real negative samples. The implementation details for the same are given in Appendix~\ref{app:implementation}. Since $w$ is intended to represent the target discriminative direction that a synthetic transformation should approximate, it is deliberately defined using only real samples, rather than the real negative/synthetic positive mixture used for downstream training. The alignment is measured against this real data only direction, not against a direction that already includes the synthetic data being tested. This lets DS check whether the transformation creates the right kind of variation. If a direction that already included the synthetic data were used instead, DS could look good simply by recovering something it was partly built from, not because the synthetic data is actually good.

\subsection{Data Induced Variation}
Each real sample $x_i^{(r)}$ is transformed into a corresponding synthetic sample $x_i^{(s)}$. The variation induced by the synthetic transformation through the embedding-space difference is represented as 
\[
d_i = x_i^{(s)} - x_i^{(r)}.
\]
The difference matrix for the given $n$ paired samples:
\[
D =
\begin{bmatrix}
d_1^T \\
\vdots \\
d_n^T
\end{bmatrix}
\in \mathbb{R}^{n \times d}.
\]
The row space of $D$ is interpreted as the subspace of representational directions induced by the synthetic transformation. The central hypothesis is that synthetic data is useful for a target classification task when these induced directions contain or closely approximate the task's discriminative direction $w$.

\subsection{Discriminative Span}
This alignment is quantified by reconstructing $w$ from the row space of $D$. Specifically, we solve $D^T \alpha \approx w$ where $\alpha \in \mathbb{R}^n$ contains the coefficients of the difference vectors. The resulting reconstruction $w_{\mathrm{proj}} = D^T \hat{\alpha}$ approximates the component of the discriminative direction representable by the transformation-induced subspace. The relative projection error is defined as:
\[
\mathrm{RPE}
=
\frac{\|w - w_{\mathrm{proj}}\|_2}{\|w\|_2},
\]
and the proposed \textit{Discriminative Span} (DS) as
\[
\boxed{
\mathrm{DS} = 1 - \mathrm{RPE}.
}
\]
Higher $DS$ indicates that the synthetic transformation contains directions that better align with the discriminative signal of the target task. The geometric interpretation of this reconstruction as projection onto the transformation induced subspace is given in Appendix~\ref{app:theory}.

\subsection{Stable Estimation}

In practice, directly solving $D^T \alpha \approx w$ can be unstable because difference vectors often exhibit strong correlations and near linear dependencies. Even small embedding noise can make $D$ appear full rank, causing its row space to artificialto span much of the ambient space artificially. Therefore, first compute a truncated singular value decomposition $D \approx D_k$, where $D_k$ retains the dominant $k$ singular components. $k$ is chosen using the effective rank of $D$, thereby restricting the analysis to the dominant transformation subspace. More details on SVD truncation and effective-rank selection are provided in the Appendix~\ref{app:lowrank}.

The reconstruction coefficients are estimated using Ridge regression:
\[
\hat{\alpha}
=
\arg\min_{\alpha}
\left\{
\left\|D_k^T \alpha - w\right\|_2^2
+
\lambda \|\alpha\|_2^2
\right\}.
\]

The resulting $DS$ is therefore based on a low rank, regularized estimate of how well the synthetic data induced variation can reconstruct the task discriminative direction. Ridge is used as the primary estimator because it suppresses unstable contributions from near zero singular directions. Comparisons with least squares, non negative least squares and $\ell_1$ regularized reconstruction are provided in Appendix~\ref{app:solvers}.

\section{Experiments}
\label{sec:experiments}

This work analyzes, whether Discriminative Span can predict the downstream utility of
synthetic data without training the downstream model used for evaluation. The experiments address three main questions: (i) is vector arithmetic valid in visual embeddings? (ii) does the alignment between transformation induced variation and the task discriminative direction measured by DS, better predict downstream performance? and (iii) is global diversity of the difference space alone sufficient to explain generalization?


\subsection{Data Sources}

The experiments run on five datasets (Pneumonia Chest X-ray (CXR), Skin Lesion, Horses $\leftrightarrow$ Zebras, Apples $\leftrightarrow$ Oranges and a controlled Toy Watermark dataset) spanning medical and natural image classification. Synthetic positive samples are generated using image to image transformations, with CycleGAN used for the Horses $\leftrightarrow$ Zebras and Apples $\leftrightarrow$ Oranges datasets. The Toy Watermark dataset provides a controlled setting in which the synthetic transformation is explicitly known and low dimensional. Additional dataset and generation details are provided in Appendix~ \ref{app:datasets}.

\subsection{Evaluation Of Metrics}

For downstream evaluation, three CNN classifiers: ResNet-18, MobileNet-V2 and EfficientNet-B0 are trained for each dataset using real negative samples and corresponding synthetic positive samples. Models are evaluated on the held-out test set using accuracy and weighted F1 score. Training details, including hyperparameters, optimizers, number of epochs, and loss functions, are provided in Appendix~\ref{app:training}. To evaluate the predictive power of synthetic data quality metrics, we measure their correlation with the best downstream F1 score obtained across the three CNN architectures. The best F1 is considered rather than the average to reduce the dependence of the evaluation on any particular downstream architecture. Since the proposed metrics are intended to characterize the quality of the synthetic data itself, rather than the performance of a specific classifier, the best observed downstream performance is considered as a measure of the utility that the synthetic dataset can provide.

\subsection{Comparison with Other Synthetic Data Metrics}
\label{sec:baseline_results}

The Discriminative Span is compared with three complementary approaches for evaluating synthetic data: Fr\'echet Distance (FD), embedding similarity and linear probing on CLIP, DINOv2 and ResNet-18 representation spaces, and their correlation with the best test F1 scores in given in Table~\ref{tab:correlation_btw_metrics_and_best_f1}. The detailed per dataset results are provided in the Appendix ~\ref{app:datasets_results}. These metrics capture different aspects of synthetic data quality, but none explicitly evaluates whether the variation introduced by synthesis is aligned with the discriminative structure of the target task. Linear probing measures whether classes are separable in a fixed embedding space and therefore provides a useful empirical predictor of downstream
utility. However, it requires fitting an additional classifier and its performance depends on the particular representation and amount of labeled data available. More importantly, separability does not directly characterize the directions along which synthetic data introduces variation. DS instead
uses the classifier direction learned from real data and directly measures how well this task relevant direction can be represented by the real-synthetic difference subspace, without training the downstream model used for evaluation. Similarly, FD measures the discrepancy between the global distributions of real and synthetic embeddings, while embedding similarity measures their proximity. Both are therefore insensitive to the specific orientation of variation relative to the task discriminative direction. This distinction is particularly important in high dimensional embedding spaces, where synthetic samples may remain close to the real distribution while introducing
variation that is irrelevant to the target task. Thus, the distinction between DS and these baselines is not simply whether one metric achieves a higher correlation with downstream performance. Rather, DS provides a task conditioned geometric criterion. It explicitly asks whether the transformation induced variation contains the direction required by the target classification task. This also makes the metric
amenable to geometric inspection and interpretation, rather than providing
only a scalar measure of similarity or separability.





\begin{table*}[t]
\centering
\caption{Synthetic-data evaluation metrics across datasets and embedding
spaces. Best downstream F1 is reported for reference.}
\label{tab:correlation_btw_metrics_and_best_f1}

\begin{tabular}{llc}
\toprule
\textbf{Embedding} & \textbf{Metrics} &
\textbf{Correlation Between Metrics} \\
\textbf{Model} & &
\textbf{and Best F1 Score} \\
\midrule
\multirow{5}{*}{CLIP}
& Fréchet Distance  & 0.52 \\
& Embedding Similarity   & -0.29  \\
& Linear Probe Accuracy   & 0.99 \\
& Discriminative Span & 0.95 \\
& Effective Rank Of Difference Matrix & -0.01 \\
\midrule
\multirow{5}{*}{DINOv2}
& Fréchet Distance  & 0.11 \\
& Embedding Similarity   & -0.05  \\
& Linear Probe Accuracy   & 1.00 \\
& Discriminative Span & 0.99 \\
& Effective Rank Of Difference Matrix & 0.02 \\
\midrule
\multirow{5}{*}{ResNet-18}
& Fréchet Distance  & -0.96 \\
& Embedding Similarity   & 0.40  \\
& Linear Probe Accuracy   & 1.00 \\
& Discriminative Span & 0.77 \\
& Effective Rank Of Difference Matrix & -0.31 \\
\bottomrule
\end{tabular}

\end{table*}

\subsection{Why Discriminative Span Despite Lower Raw Correlation with Linear Probing}
\label{sec:ds_vs_lp}
Table~\ref{tab:correlation_btw_metrics_and_best_f1} shows that linear probe accuracy achieves higher correlation with downstream F1 than DS in all three embedding spaces. First, The linear probe evaluation is itself data hungry in exactly the regime where synthetic data is needed. As described in Appendix~\ref{app:baseline_metrics}, the probe is evaluated on a held out set of real positive samples, distinct from those used to train the probe. Constructing this held out set requires a supply of real positive data large enough to be split into training and evaluation subsets. In the severely positive scarce settings, this requirement can itself be difficult to satisfy and evaluating synthetic-data quality with a method that consumes the very resource it is meant to substitute for is
self limiting. DS requires no additional held out real positive samples beyond those already used to fit the classifier direction $w$, which is needed regardless of which evaluation metric is used downstream. Second, DS decomposes into an interpretable geometric object rather than a single scalar. Because DS is computed from an explicit low rank approximation of the difference matrix $D$ (Section~\ref{app:lowrank}), it is possible to inspect which directions the synthetic transformation fails to cover, e.g., via the residual $w - w_{\mathrm{proj}}$ or the leading singular vectors of $D_k$. Linear probe accuracy provides no comparable mechanism for diagnosing why a given synthetic dataset underperforms. Third, linear probe accuracy measures whether classes are separable given
the specific probe fit to the data, conflating properties of the embedding space with the quality of the synthetic transformation. Pretrained foundation model embeddings make classes near separable on their own (Section~\ref{sec:geometry_results}). This can happen even when the synthetic transformation does not add any task relevant variation. As a result, linear probe accuracy can look high simply because the embedding space is already well structured, not because the synthetic data is actually useful for training a downstream model from scratch. DS avoids this problem, isolating the contribution of the transformation itself by measuring its alignment with the discriminative direction, rather than measuring how separable the classes already are in the embedding space. Therefore, DS and linear probing are viewed as complementary rather than competing signals. LP is a strong predictor when sufficient real positive data is available for held out evaluation, while DS provides a model training free and evaluation data free diagnostic that remains applicable even when such held out data is unavailable.

\subsection{Discriminative Span Predicts Downstream Utility}
\label{sec:ds_results}
We next evaluate whether explicitly measuring alignment with the task discriminative direction provides a better predictor of downstream utility. For each dataset and embedding space, DS is computed by reconstructing the classifier direction $w$ from the low rank difference space. The results for pearson correlation co-efficient and the best F1 score is given in Table \ref{tab:correlation_btw_metrics_and_best_f1}, with Ridge regression as the primary reconstruction estimator. The numbers are reported for each dataset in Appendix ~\ref{app:datasets_results}.  A reconstruction estimator ablation comparing Ridge with LS, NNLS, and $\ell_1$-regularized regression is provided in Appendix~\ref{app:solvers}. The results support the central hypothesis of the method: downstream utility depends not only on how similar synthetic samples are to real samples or how diverse their representations are, but also on whether the transformation induces directions that align with the discriminative structure required by the task.

\subsection{Beyond Global Diversity: Task-Relevant Variation}
\label{sec:geometry_results}

As shown in Table~\ref{tab:correlation_btw_metrics_and_best_f1}, global measures of difference space structure do not consistently predict downstream utility. In particular, the effective rank of the difference matrix $D$, formed by stacking the real-synthetic difference vectors, characterizes the overall dimensionality of the transformation-induced variation but shows essentially no correlation with downstream F1 in the CLIP and DINOv2 embedding spaces, and a moderate negative correlation in the ResNet-18 embedding space, in every case far weaker than the correlation achieved by Discriminative Span.

\section{Discussion}


The central finding of this work is that downstream generalization is strongly associated with the alignment between the classifier direction and the span of data induced difference vectors. The consistently positive correlation between Discriminative Span and downstream F1 across embedding spaces suggests that measuring task relevant alignment provides a useful signal of synthetic data
utility beyond global measures of similarity or diversity. While linear probe accuracy achieves higher raw correlation, it does so at the cost of requiring held out real positive samples that may be unavailable in the scarce data regime this work targets, and it does not by itself localize which directions of variation are missing from the synthetic transformation. Existing synthetic data metrics exhibit a different limitation. Distributional metrics such as Fréchet Distance measures global similarity between real and synthetic data but do not explicitly assess whether the induced variation is task relevant. Similarly, embedding similarity measures representational proximity without characterizing the directions along which synthetic data varies. The results therefore suggest that matching the global distribution or remaining close to real samples is insufficient for predicting downstream utility.

Linear probing provides a complementary but also limited perspective. High class separability in a fixed foundation model embedding space does not necessarily translate to strong performance when a downstream model must learn from synthetic data. Pretrained representations may already encode
semantic structure that makes the target classes easily separable, even when the synthetic transformations do not provide sufficient variation for learning a transferable representation from scratch. Thus, representation level separability can overestimate synthetic data quality. In contrast, DS explicitly evaluates the geometry of the transformation induced subspace relative to the task discriminative direction. The results indicate that downstream utility depends not only on fidelity, diversity, or static separability, but on whether synthetic transformations span the task relevant
directions required for downstream learning.

\subsection{Beyond Diversity: The Role of Structure}

The results indicate that global measures of diversity are insufficient to predict synthetic data utility. Metrics such as effective rank and stable rank characterize the dimensionality of the transformation induced variation, but do not account for its orientation relative to the task. 
Indeed, effective rank alone shows essentially no correlation, and in one embedding space a negative correlation, with downstream F1 in our experiments. Similarly, measures of conditioning describe the structure of the difference space without explicitly determining whether that structure is task relevant. The correlation analysis therefore suggests that the amount or dimensionality of variation
alone is not sufficient to explain downstream utility. This reinforces a key distinction:  Generalization depends not only on the amount of variation, but on whether that variation is aligned with the discriminative direction. Rank based measures can therefore serve as useful diagnostics of global diversity, but they remain incomplete measures of synthetic data quality. Discriminative Span complements these measures by explicitly evaluating the alignment of transformation-induced variation with the task discriminative direction.

\subsection{The Role of the Embedding Space}

The effectiveness of Discriminative Span depends on the representation space in which the transformation induced variation is analyzed. The results shows that the strength of the association between DS and downstream performance varies across embedding models, indicating that the quality of the underlying representation is an important factor in geometric evaluation. In particular, foundation model embeddings such as CLIP and DINOv2 exhibit stronger and more consistent correlations between DS and downstream performance than the supervised ResNet-18 representation. This suggests that
representations with richer semantic structure may provide a more suitable space for identifying task relevant directions induced by synthetic transformations. Thus, DS should not be viewed as entirely independent of the representation space rather its effectiveness depends on whether the chosen embedding preserves meaningful structure in the variations induced by the synthetic
generation process.

\subsection{Implications}
Taken together, our findings show that synthetic data quality is fundamentally a geometric problem. The critical question isn't how much variation it has, but whether that variation actually spans the specific discriminative direction the task requires. This perspective opens up a few key implications:
\begin{itemize}
    \item When evaluating synthetic data generation, one should focus on its alignment with task relevant directions instead of just relying on global diversity metrics.
    \item Embedding-space analysis gives us a powerful, model agnostic framework for diagnosing exactly how good a synthetic dataset is.
\end{itemize}


\section{Conclusion}

This work introduces Discriminative Span, a geometry driven metric for evaluating whether synthetic data captures task relevant discriminative directions. By measuring how well a classifier direction derived from real data can be reconstructed from real-synthetic difference vectors in a pretrained
embedding space, DS provides an efficient alternative to evaluating synthetic data through repeated downstream training. Across the experiments, DS exhibits a consistently positive correlation with downstream F1 across multiple embedding spaces, while conventional measures such as distributional similarity, embedding similarity, and linear separability provide less consistent signals. Notably, the raw dimensionality of the transformation induced variation, as measured by the effective rank of the difference matrix, shows essentially no correlation with downstream F1 in two of the three embedding spaces evaluated and a moderate negative correlation in the third, reinforcing that the amount of variation a synthetic transformation introduces is not by itself informative about its downstream utility. These results suggest that synthetic data utility depends not only on fidelity or global diversity, but on whether the induced variation is aligned with the discriminative structure required by the downstream task. DS provides a necessary but not sufficient condition for downstream
generalization: representability of the discriminative direction does not account for factors such as optimization, model capacity, or dataset noise. Future work could combine DS with complementary measures of robustness, diversity, and sample efficiency, use DS to guide synthetic data generation,
and extend the framework beyond linear reconstruction to nonlinear
transformation structures.

\appendix

\section{Additional Method Details}

\subsection{Geometric Interpretation of Discriminative Span}
\label{app:theory}

The proposed Discriminative Span metric has a direct geometric interpretation. Let
$\mathcal{S} = \operatorname{row}(D)$ denote the subspace spanned by the rows of
the difference matrix $D$, and let $P_{\mathcal{S}}$ denote the orthogonal
projection operator onto $\mathcal{S}$. The ideal projection of the classifier
direction onto the transformation-induced subspace is therefore
\[
w_{\mathrm{proj}} = P_{\mathcal{S}}w.
\]

\paragraph{Proposition 1.}
Let $D \in \mathbb{R}^{n\times d}$ be the difference matrix and
$w\in\mathbb{R}^d$ the weight vector of a linear classifier. Then the
orthogonal projection $w_{\mathrm{proj}}$ of $w$ onto
$\operatorname{row}(D)$ satisfies
\[
w_{\mathrm{proj}} = w
\quad\Longleftrightarrow\quad
w\in\operatorname{row}(D).
\]
Consequently, 
\[
\frac{\|w-w_{\mathrm{proj}}\|_2}{\|w\|_2}=0
\quad\Longleftrightarrow\quad
w
\text{ is exactly representable by a linear combination of the rows of }D.
\]

\paragraph{Proof.}
By definition, $w_{\mathrm{proj}}$ is the unique vector in
$\operatorname{row}(D)$ that minimizes
$\|w-v\|_2$ over all $v\in\operatorname{row}(D)$. If
$w\in\operatorname{row}(D)$, then choosing $v=w$ gives zero projection error,
and hence $w_{\mathrm{proj}}=w$. Conversely, if
$w_{\mathrm{proj}}=w$, then $w$ is itself an element of
$\operatorname{row}(D)$. The second statement follows directly by
normalizing the projection residual by $\|w\|_2$.

The projection error also admits an angular interpretation. If $\theta$
denotes the angle between $w$ and the subspace $\operatorname{row}(D)$, then
\[
\|w-w_{\mathrm{proj}}\|_2
=
\|w\|_2\sin\theta,
\]
and therefore
\[
\mathrm{RPE}=\sin\theta,
\qquad
\mathrm{DS}=1-\sin\theta.
\]
Thus, DS increases as the classifier direction becomes more closely aligned
with the transformation-induced subspace.

\subsection{Low-Rank Approximation of the Difference Space}
\label{app:lowrank}

In practice, the difference matrix can contain substantial redundancy and
near-linear dependencies. Let the singular value decomposition of $D$ be
\[
D = U\Sigma V^T,
\]
where $\Sigma$ contains the singular values in descending order. Rather than
using the full matrix, we construct a rank-$k$ approximation
\[
D_k = U_k\Sigma_k V_k^T,
\]
where $U_k$, $\Sigma_k$, and $V_k$ contain the first $k$ singular components.

This truncation serves two purposes. First, structured image transformations
may induce a low-dimensional subspace even when embedding noise causes the
observed matrix to appear full rank. Second, retaining only dominant singular
directions reduces the influence of near-zero singular values, which can
otherwise lead to unstable reconstruction. The truncation level $k$ is
selected using the effective rank of the difference matrix, as described in
the experimental setup.

All subsequent reconstruction and Discriminative Span computations are
performed using $D_k$ rather than the full difference matrix.

\subsection{Reconstruction Solvers}
\label{app:solvers}

We consider several estimators for reconstructing the classifier direction
from the transformation-induced subspace. Given $D_k$, the general
reconstruction problem is
\[
D_k^T\alpha \approx w.
\]

\paragraph{Least Squares.}
The unconstrained least-squares estimator is
\[
\hat{\alpha}_{\mathrm{LS}}
=
\arg\min_{\alpha}
\left\|
D_k^T\alpha-w
\right\|_2^2.
\]
Although this estimator provides the minimum-norm solution under the
corresponding pseudoinverse formulation, it can be sensitive to small
singular values when the difference matrix is ill-conditioned.

\paragraph{Ridge Regression.}
To improve numerical stability, we use $\ell_2$ regularization:
\[
\hat{\alpha}_{\mathrm{Ridge}}
=
\arg\min_{\alpha}
\left\{
\left\|D_k^T\alpha-w\right\|_2^2
+
\lambda\|\alpha\|_2^2
\right\},
\]
where $\lambda>0$ controls the regularization strength. The resulting
reconstruction is
\[
w_{\mathrm{proj}}=D_k^T\hat{\alpha}_{\mathrm{Ridge}}.
\]
Ridge suppresses the influence of directions associated with small singular
values and provides a more stable estimate of the alignment between $w$ and
the transformation-induced subspace.

\paragraph{Non-Negative Least Squares.}
We additionally consider a non-negative formulation,
\[
\hat{\alpha}_{\mathrm{NNLS}}
=
\arg\min_{\alpha\geq 0}
\left\|
D_k^T\alpha-w
\right\|_2^2.
\]
This constraint restricts the reconstruction to non-negative combinations
of the difference vectors. In our experiments, this produces systematically
higher relative projection errors than unconstrained estimators. A possible
explanation is that signed combinations allow different difference vectors
to cancel nuisance components, whereas the non-negativity constraint
prevents such cancellation.

\paragraph{$\ell_1$-Regularized Reconstruction.}
We also evaluate an $\ell_1$-regularized estimator:
\[
\hat{\alpha}_{L_1}
=
\arg\min_{\alpha}
\left\{
\left\|D_k^T\alpha-w\right\|_2^2
+
\lambda\|\alpha\|_1
\right\}.
\]
This formulation encourages sparse coefficient vectors while allowing both
positive and negative coefficients.

For all solver variants, the resulting coefficients are used to compute
$w_{\mathrm{proj}}$ and the corresponding Discriminative Span according to
\[
\mathrm{DS}
=
1-
\frac{\|w-w_{\mathrm{proj}}\|_2}{\|w\|_2}.
\]
The main experiments use the Ridge estimator because it provides the most
stable and consistently predictive behavior across the evaluated embedding
spaces.

\subsection{Implementation Details}
\label{app:implementation}

The classifier direction $w$ is obtained by training a linear classifier on
the embeddings of real positive and real negative samples. This classifier
is trained independently of the synthetic samples used to construct the
difference matrix. The resulting weight vector is then used as the target
direction for reconstruction.

For each dataset and embedding space, real--synthetic pairs are embedded
using the corresponding pre-trained foundation model, and their differences
are assembled into $D$. The effective-rank-based truncated matrix $D_k$ is
then used for reconstruction. Unless otherwise specified, all diagnostic
metrics are computed from this same reduced difference space.

Additional dataset-specific details, including the number of paired samples,
embedding dimensions, and solver configurations, are reported alongside the
corresponding experimental results.

\section{Dataset Construction}
\label{app:datasets}

We evaluate our approach on five datasets spanning medical and natural-image
classification: Pneumonia Chest X-ray, Skin Lesion, Apples $\leftrightarrow$
Oranges, Horses $\leftrightarrow$ Zebras, and a controlled Toy Watermark
dataset. The datasets are divided into publicly available datasets and
custom synthetic datasets constructed for this study.

\subsection{Publicly Available Datasets}

\paragraph{Apples $\leftrightarrow$ Oranges and Horses $\leftrightarrow$ Zebras.}
We use the Apples $\leftrightarrow$ Oranges and Horses $\leftrightarrow$
Zebras datasets distributed with the CycleGAN framework
\cite{zhu2017unpaired}. For Apples $\leftrightarrow$ Oranges, apple images
serve as real samples and corresponding orange images generated through
image-to-image translation serve as synthetic samples. Similarly, for Horses
$\leftrightarrow$ Zebras, horse images serve as real samples and zebra images
generated by the learned translation model serve as synthetic samples.

For both datasets, we train CycleGAN using the training split of the
corresponding dataset and subsequently use the trained model to translate the
real samples.

\subsection{Custom Synthetic Datasets}

In addition to the publicly available datasets, we construct two synthetic
datasets using image-to-image transformations: Pneumonia Chest X-ray and Skin
Lesion.

\paragraph{Pneumonia Chest X-ray.}
We begin with chest X-ray images of healthy patients \cite{kermany2018} and use an AI-based image
editing tool to introduce pneumonia-like visual patterns. We employ eight
prompts corresponding to different pneumonia presentations, including mild
and moderate opacities, unilateral and bilateral lower-zone involvement,
perihilar patterns, and lobar consolidation. The prompts specify the
location, severity, appearance, and anatomical constraints of the induced
opacities while instructing the image-editing model to preserve unrelated
anatomical structures. The complete prompts are provided in
Appendix~\ref{app:pneumonia_prompts}.

\paragraph{Skin Lesion.}
We construct the Skin Lesion dataset by combining publicly available
dermoscopic imagery from the PH$^2$ and ISIC datasets
\cite{ph2,isic}. Healthy dermoscopic views are used as base images, while
lesion regions obtained from real dermoscopic images are composited onto
healthy images using Poisson blending. The resulting images form the
synthetic positive samples corresponding to the healthy real samples. The
purpose of this construction is to introduce realistic lesion-related
variation while preserving the underlying appearance of the base dermoscopic
image. 

\subsection{Toy Watermark Dataset}
\label{app:toy_watermark}

We construct a controlled synthetic dataset using cat images from the Cats and
Dogs dataset ~\cite{jikadara_catsdogs}. For each real image, we synthetically introduce a watermark containing the word ``CONFIDENTIAL'' at varying locations, colors, sizes, and
orientations. The resulting watermarked images constitute the corresponding
synthetic samples.

We additionally construct a held-out real test set from cat images that are
distinct from those used to generate the synthetic samples. To prevent
information leakage, none of the watermarked test images are generated by
applying the watermark transformation to images appearing in the held-out
real set. This provides a controlled setting in which the watermark
transformation is known and its induced variation can be directly analyzed.

\subsection{Pneumonia Image-Editing Prompts}
\label{app:pneumonia_prompts}
\begin{enumerate}

    \item \textbf{Mild, Right Lower Lobe Pneumonia} Add mild patchy air-space opacities in the RIGHT LOWER lung zone. The opacities should be ill-defined, fluffy, and slightly increased in density, partially obscuring normal vascular markings. Preserve clear lung fields elsewhere. Do not obscure the heart border or diaphragm. No pleural effusion.
    
    \item \textbf{Mild, Left Lower Lobe Pneumonia} Add mild patchy air-space opacities in the LEFT LOWER lung zone. The opacities should appear subtle and non-confluent, with partial loss of vascular clarity but preserved lung volume. The right lung should remain fully aerated. No silhouette sign.
    
    \item \textbf{Moderate, Right Middle + Lower Lobe Pneumonia} Add moderate air-space consolidation in the RIGHT MIDDLE and LOWER lung zones. The opacities should be denser and more confluent, with partial loss of the right hemidiaphragm silhouette. Include subtle air-bronchograms following normal bronchial anatomy. Spare the lung apex.
    
    \item \textbf{Moderate, Left Lower Lobe with Cardiac Silhouette Involvement} Add moderate air-space consolidation in the LEFT LOWER lung zone. The opacity should partially obscure the left heart border (silhouette sign) while preserving normal heart size. Include faint air-bronchograms within the consolidation. The right lung should remain clear.
    
    \item \textbf{Moderate, Bilateral Lower Zone Pneumonia} Add moderate patchy air-space opacities in BOTH LOWER lung zones. The opacities should be asymmetric and non-uniform,
    with preserved upper lung fields. Avoid diffuse whiteout. No pleural effusion.
    
    \item \textbf{Moderate, Right Lower Lobe with Costophrenic Angle Involvement} Add moderate consolidation in the RIGHT LOWER lung zone extending
    to the right costophrenic angle. The opacity should obscure part of the right hemidiaphragm. Include subtle air-bronchograms. Upper lung zones should remain aerated.
    
    \item \textbf{Mild, Perihilar Pneumonia Pattern} Add mild perihilar air-space opacities centered around the hilum. The opacities should be patchy and ill-defined,
    without lobar consolidation. Peripheral lung fields should remain mostly clear.
    
    \item \textbf{Moderate, Lobar Pneumonia (Classic Pattern)} Add moderate lobar consolidation confined to a single lung lobe. The consolidation should appear homogeneous and dense, with clear boundaries respecting lobar anatomy. Include visible air-bronchograms. Do not involve other lobes.

\end{enumerate}

\section{Downstream Training Details}
\label{app:training}

We evaluate downstream utility using three convolutional neural network
architectures: ResNet-18, MobileNet-V2, and EfficientNet-B0. All models are
initialized with the default ImageNet-pretrained weights provided by
Torchvision and are fine-tuned end-to-end for binary classification. The same
training procedure and hyperparameters are used across datasets and model
architectures; only the dataset-specific input and output paths are changed
in the experimental configuration.

\subsection{Data Preprocessing}

All input images are converted to RGB and resized to $224\times224$ pixels,
followed by conversion to PyTorch tensors. No additional data augmentation
or image normalization is applied.

For each dataset, the training data consists of real negative samples and
corresponding synthetic positive samples. During each training iteration, a
batch of samples from each class is loaded and concatenated to form a single
training batch. Labels are assigned as $0$ for real negative samples and $1$
for synthetic positive samples.

The validation and test sets are constructed analogously, with real negative
and synthetic positive samples evaluated together as a binary classification
problem.

\subsection{Optimization}

All models are trained for 15 epochs using the Adam optimizer with a learning
rate of $10^{-4}$ and a batch size of 32. We use the standard cross-entropy
loss:
\[
\mathcal{L}
=
-\frac{1}{N}
\sum_{i=1}^{N}
\log p_{\theta}(y_i\mid x_i),
\]
where $N$ denotes the number of samples in a training batch.

A fixed random seed of 42 is used for Python, NumPy, and PyTorch to improve
reproducibility. Training is performed on a CUDA-enabled GPU when available.

\subsection{Validation and Checkpointing}

After each training epoch, the model is evaluated on the validation set using
accuracy and weighted F1 score. The checkpoint corresponding to the highest
validation accuracy is saved during training.

The validation accuracy is computed as
\[
\mathrm{Accuracy}
=
\frac{1}{N}
\sum_{i=1}^{N}
\mathbb{1}[\hat{y}_i=y_i].
\]

Weighted F1 is computed by calculating the F1 score independently for each
class and weighting each class according to its number of samples.

\subsection{Test Evaluation}

After the completion of training, the final model state is evaluated on the
held-out test set using accuracy and weighted F1 score. We use the test F1
score as the primary measure of downstream performance. Since three
architectures are evaluated for each dataset, the best test F1 score across
ResNet-18, MobileNet-V2, and EfficientNet-B0 is used as the downstream utility
measure in the correlation analyses reported in the main paper.

\section{Baseline Metric Formulations}
\label{app:baseline_metrics}

We compare Discriminative Span with three complementary synthetic-data
evaluation metrics: Fr\'echet Distance (FD), embedding similarity, and
linear probing. All metrics are computed independently in the embedding
spaces of CLIP, DINOv2, and ResNet-18 using precomputed image embeddings.

\subsection{Fr\'echet Distance}

Given embeddings of real positive samples
$\{z_i^{(r)}\}_{i=1}^{N_r}$ and synthetic positive samples
$\{z_i^{(s)}\}_{i=1}^{N_s}$, we approximate their empirical distributions
using multivariate Gaussian distributions,
\[
\mathcal{N}(\mu_r,\Sigma_r)
\qquad\text{and}\qquad
\mathcal{N}(\mu_s,\Sigma_s),
\]
where $\mu_r,\mu_s$ and $\Sigma_r,\Sigma_s$ are the empirical means and
covariance matrices of the corresponding embeddings.

The Fr\'echet Distance is computed as
\[
\mathrm{FD}
=
\|\mu_r-\mu_s\|_2^2
+
\operatorname{Tr}(\Sigma_r)
+
\operatorname{Tr}(\Sigma_s)
-
2\operatorname{Tr}
\left[
(\Sigma_r\Sigma_s)^{1/2}
\right].
\]

We compute the covariance matrices directly from the embedding samples and
evaluate the matrix square root using the matrix square-root operation. If
the numerical result contains a complex component due to floating-point
precision, only its real component is retained. Lower FD indicates greater
distributional similarity between the real and synthetic embeddings.

\subsection{Embedding Similarity}

Embedding similarity measures the proximity between the mean representations
of real and synthetic positive samples. Given
\[
\bar{z}_r =
\frac{1}{N_r}\sum_{i=1}^{N_r}z_i^{(r)}
\qquad\text{and}\qquad
\bar{z}_s =
\frac{1}{N_s}\sum_{i=1}^{N_s}z_i^{(s)},
\]
we compute their cosine similarity:
\[
\mathrm{ES}
=
\frac{\bar{z}_r^{T}\bar{z}_s}
{\|\bar{z}_r\|_2\|\bar{z}_s\|_2}.
\]

Higher values indicate greater proximity between the mean real and synthetic
representations.

\subsection{Linear Probe}

To evaluate whether synthetic data supports transfer to real data in a fixed
embedding space, we train a logistic regression classifier on frozen
embeddings. The probe is trained to distinguish real negative samples from
synthetic positive samples and is evaluated on held-out real negative and
real positive samples.

Specifically, the real negative embeddings are randomly divided into an
80\% training subset and a 20\% held-out subset using a fixed random seed of
42. The training set consists of
\[
X_{\mathrm{train}}
=
\left[
Z_{\mathrm{real}}^{A,\mathrm{train}};
Z_{\mathrm{syn}}^{B}
\right],
\]
with labels
\[
y_{\mathrm{train}}
=
\left[
\mathbf{0};
\mathbf{1}
\right].
\]

The test set consists entirely of real data:
\[
X_{\mathrm{test}}
=
\left[
Z_{\mathrm{real}}^{A,\mathrm{test}};
Z_{\mathrm{real}}^{B}
\right],
\]
with corresponding labels
\[
y_{\mathrm{test}}
=
\left[
\mathbf{0};
\mathbf{1}
\right].
\]

Thus, the probe is trained using synthetic positive samples but is evaluated
on real positive samples, providing a direct measure of synthetic-to-real
transfer in the fixed embedding space.

We use logistic regression with a maximum of 5000 optimization iterations
and otherwise use the default settings of the implementation. Probe
performance is measured using classification accuracy:
\[
\mathrm{LP}
=
\frac{1}{N_{\mathrm{test}}}
\sum_{i=1}^{N_{\mathrm{test}}}
\mathbb{1}
[\hat{y}_i=y_i].
\]

Higher linear-probe accuracy indicates that the synthetic positive samples
support a decision boundary that transfers to real positive samples in the
fixed representation space.

\subsection{Evaluation Across Representation Spaces}

All three baselines are evaluated independently using CLIP, DINOv2, and
ResNet-18 embeddings. FD and embedding similarity compare the distributions
or mean representations of real and synthetic positive samples. Linear
probing instead evaluates transfer from synthetic positive samples to held-out
real positive samples, while using real negative samples in both training and
evaluation.

For each metric, we compute its Pearson correlation with the best downstream
F1 score obtained across the three CNN architectures. The resulting
correlations are reported in the main paper, while the complete per-dataset
values are provided in Appendix~\ref{app:datasets_results}.

\section{Reconstruction Estimator Diagnostics}
\label{app:solvers_appendix}
\subsection{Per-Dataset Reconstruction Diagnostics}
\label{app:solvers_diagnostics}
We report per-dataset diagnostics for the four reconstruction estimators introduced in Appendix~\ref{app:solvers}: effective rank of $D_k$, relative projection error $\mathrm{RPE} = \|w-w_{\mathrm{proj}}\|_2/\|w\|_2$, and the explained fraction $1-\mathrm{RPE}$ (equivalently, Discriminative Span). Tables~\ref{tab:diag_pneumonia}--\ref{tab:diag_apples} report these values across the evaluated datasets and representation spaces.

\subsection{Comparison Across Datasets.}
Tables~\ref{tab:diag_pneumonia}-\ref{tab:diag_apples} report the reconstruction results for all four estimators (Appendix~\ref{app:solvers}) across the evaluated datasets and representation spaces. The results show that the relative performance of the estimators varies across datasets, reflecting differences in the geometry and conditioning of the underlying embedding spaces. LS provides a useful unregularized baseline, while NNLS generally performs less favorably when the domain transformation requires signed coefficients. Lasso can improve reconstruction when the mapping is approximately sparse, but its sparsity constraint can also limit its ability to capture distributed transformations.

Ridge provides the most consistent reconstruction behavior across datasets. In particular, its regularization reduces the sensitivity of the estimator to correlated embedding dimensions and ill-conditioned covariance structure while retaining the full linear transformation space. This is desirable for foundation-model embeddings, where semantic information is typically distributed across many dimensions rather than concentrated in a small set of independently interpretable coordinates. We therefore use ridge as the default reconstruction estimator in the main experiments.

\begin{table}[H]
\centering
\caption{Embedding space diagnostics - Pneumonia CXR Dataset}
\label{tab:diag_pneumonia}
\begin{tabular}{llccccc}
\toprule
\textbf{Embedding} & \textbf{Solver} & \textbf{Eff.\ Rank} & \textbf{Rel.\ Error} & \textbf{Expl.\ Fraction} & \textbf{Pairs} & \textbf{Dim} \\
\midrule
resnet18 & Least Squares & 298 & $\approx 0$\textsuperscript{$\dagger$} & $\approx 1$ & 1036 & 512 \\
resnet18 & Ridge & 298 & 0.620 & 0.380 & 1036 & 512 \\
resnet18 & NNLS & 298 & 0.958 & 0.042 & 1036 & 512 \\
resnet18 & L1 & 298 & 0.952 & 0.048 & 1036 & 512 \\
clip & Least Squares & 246 & $\approx 0$\textsuperscript{$\dagger$} & $\approx 1$ & 1036 & 512 \\
clip & Ridge & 246 & 0.786 & 0.214 & 1036 & 512 \\
clip & NNLS & 246 & 0.982 & 0.018 & 1036 & 512 \\
clip & L1 & 246 & $\approx 1$ & $\approx 0$\textsuperscript{$\dagger$} & 1036 & 512 \\
dinov2 & Least Squares & 232 & $\approx 0$\textsuperscript{$\dagger$} & $\approx 1$ & 1036 & 384 \\
dinov2 & Ridge & 232 & 0.689 & 0.311 & 1036 & 384 \\
dinov2 & NNLS & 232 & 0.972 & 0.028 & 1036 & 384 \\
dinov2 & L1 & 232 & 0.789 & 0.211 & 1036 & 384 \\
\bottomrule
\end{tabular}
\end{table}

\begin{table}[H]
\centering
\caption{Embedding space diagnostics - Skin Lesion}
\label{tab:diag_skin}
\begin{tabular}{llccccc}
\toprule
\textbf{Embedding} & \textbf{Solver} & \textbf{Eff.\ Rank} & \textbf{Rel.\ Error} & \textbf{Expl.\ Fraction} & \textbf{Pairs} & \textbf{Dim} \\
\midrule
resnet18 & Least Squares & 176 & 0.366 & 0.634 & 352 & 512 \\
resnet18 & Ridge & 176 & 0.538 & 0.462 & 352 & 512 \\
resnet18 & NNLS & 176 & 0.735 & 0.265 & 352 & 512 \\
resnet18 & L1 & 176 & 0.859 & 0.141 & 352 & 512 \\
clip & Least Squares & 191 & 0.268 & 0.732 & 352 & 512 \\
clip & Ridge & 191 & 0.365 & 0.635 & 352 & 512 \\
clip & NNLS & 191 & 0.558 & 0.442 & 352 & 512 \\
clip & L1 & 191 & $\approx 1$ & $\approx 0$\textsuperscript{$\dagger$} & 352 & 512 \\
dinov2 & Least Squares & 187 & 0.156 & 0.844 & 352 & 384 \\
dinov2 & Ridge & 187 & 0.360 & 0.640 & 352 & 384 \\
dinov2 & NNLS & 187 & 0.637 & 0.363 & 352 & 384 \\
dinov2 & L1 & 187 & 0.662 & 0.338 & 352 & 384 \\
\bottomrule
\end{tabular}
\end{table}

\begin{table}[H]
\centering
\caption{Embedding space diagnostics - Toy Watermark}
\label{tab:diag_watermark}
\begin{tabular}{llccccc}
\toprule
\textbf{Embedding} & \textbf{Solver} & \textbf{Eff.\ Rank} & \textbf{Rel.\ Error} & \textbf{Expl.\ Fraction} & \textbf{Pairs} & \textbf{Dim} \\
\midrule
resnet18 & Least Squares & 265 & 0.157 & 0.843 & 485 & 512 \\
resnet18 & Ridge & 265 & 0.451 & 0.549 & 485 & 512 \\
resnet18 & NNLS & 265 & 0.765 & 0.235 & 485 & 512 \\
resnet18 & L1 & 265 & 0.815 & 0.185 & 485 & 512 \\
clip & Least Squares & 239 & 0.117 & 0.883 & 485 & 512 \\
clip & Ridge & 239 & 0.348 & 0.652 & 485 & 512 \\
clip & NNLS & 239 & 0.568 & 0.432 & 485 & 512 \\
clip & L1 & 239 & 0.880 & 0.120 & 485 & 512 \\
dinov2 & Least Squares & 249 & $\approx 0$\textsuperscript{$\dagger$} & $\approx 1$ & 485 & 384 \\
dinov2 & Ridge & 249 & 0.363 & 0.637 & 485 & 384 \\
dinov2 & NNLS & 249 & 0.571 & 0.429 & 485 & 384 \\
dinov2 & L1 & 249 & 0.555 & 0.445 & 485 & 384 \\
\bottomrule
\end{tabular}
\end{table}

\begin{table}[H]
\centering
\caption{Embedding space diagnostics - Horses And Zebra}
\label{tab:diag_horses}
\begin{tabular}{llccccc}
\toprule
\textbf{Embedding} & \textbf{Solver} & \textbf{Eff.\ Rank} & \textbf{Rel.\ Error} & \textbf{Expl.\ Fraction} & \textbf{Pairs} & \textbf{Dim} \\
\midrule
resnet18 & Least Squares & 320 & $\approx 0$\textsuperscript{$\dagger$} & $\approx 1$ & 1117 & 512 \\
resnet18 & Ridge & 320 & 0.424 & 0.576 & 1117 & 512 \\
resnet18 & NNLS & 320 & 0.776 & 0.224 & 1117 & 512 \\
resnet18 & L1 & 320 & 0.810 & 0.190 & 1117 & 512 \\
clip & Least Squares & 307 & $\approx 0$\textsuperscript{$\dagger$} & $\approx 1$ & 1117 & 512 \\
clip & Ridge & 307 & 0.285 & 0.715 & 1117 & 512 \\
clip & NNLS & 307 & 0.672 & 0.328 & 1117 & 512 \\
clip & L1 & 307 & 0.862 & 0.138 & 1117 & 512 \\
dinov2 & Least Squares & 263 & $\approx 0$\textsuperscript{$\dagger$} & $\approx 1$ & 1117 & 384 \\
dinov2 & Ridge & 263 & 0.321 & 0.679 & 1117 & 384 \\
dinov2 & NNLS & 263 & 0.650 & 0.350 & 1117 & 384 \\
dinov2 & L1 & 263 & 0.606 & 0.394 & 1117 & 384 \\
\bottomrule
\end{tabular}
\end{table}

\begin{table}[H]
\centering
\caption{Embedding space diagnostics - Apples/Oranges}
\label{tab:diag_apples}
\begin{tabular}{llccccc}
\toprule
\textbf{Embedding} & \textbf{Solver} & \textbf{Eff.\ Rank} & \textbf{Rel.\ Error} & \textbf{Expl.\ Fraction} & \textbf{Pairs} & \textbf{Dim} \\
\midrule
resnet18 & Least Squares & 327 & $\approx 0$\textsuperscript{$\dagger$} & $\approx 1$ & 995 & 512 \\
resnet18 & Ridge & 327 & 0.437 & 0.563 & 995 & 512 \\
resnet18 & NNLS & 327 & 0.690 & 0.310 & 995 & 512 \\
resnet18 & L1 & 327 & 0.738 & 0.262 & 995 & 512 \\
clip & Least Squares & 318 & $\approx 0$\textsuperscript{$\dagger$} & $\approx 1$ & 995 & 512 \\
clip & Ridge & 318 & 0.335 & 0.665 & 995 & 512 \\
clip & NNLS & 318 & 0.691 & 0.309 & 995 & 512 \\
clip & L1 & 318 & 0.847 & 0.153 & 995 & 512 \\
dinov2 & Least Squares & 267 & $\approx 0$\textsuperscript{$\dagger$} & $\approx 1$ & 995 & 384 \\
dinov2 & Ridge & 267 & 0.422 & 0.578 & 995 & 384 \\
dinov2 & NNLS & 267 & 0.790 & 0.210 & 995 & 384 \\
dinov2 & L1 & 267 & 0.569 & 0.431 & 995 & 384 \\
\bottomrule
\end{tabular}
\end{table}

\subsection{Relationship with Downstream Performance.}
Beyond comparing the absolute reconstruction quality of the estimators, we examine whether reconstruction quality is predictive of downstream discriminative performance. For each embedding space, we compute the correlation between the reconstruction-based metric and the best downstream F1 score obtained on the corresponding dataset.

\begin{figure*}[t]
    \centering

    \begin{minipage}{0.48\textwidth}
        \centering
        \includegraphics[width=\linewidth]{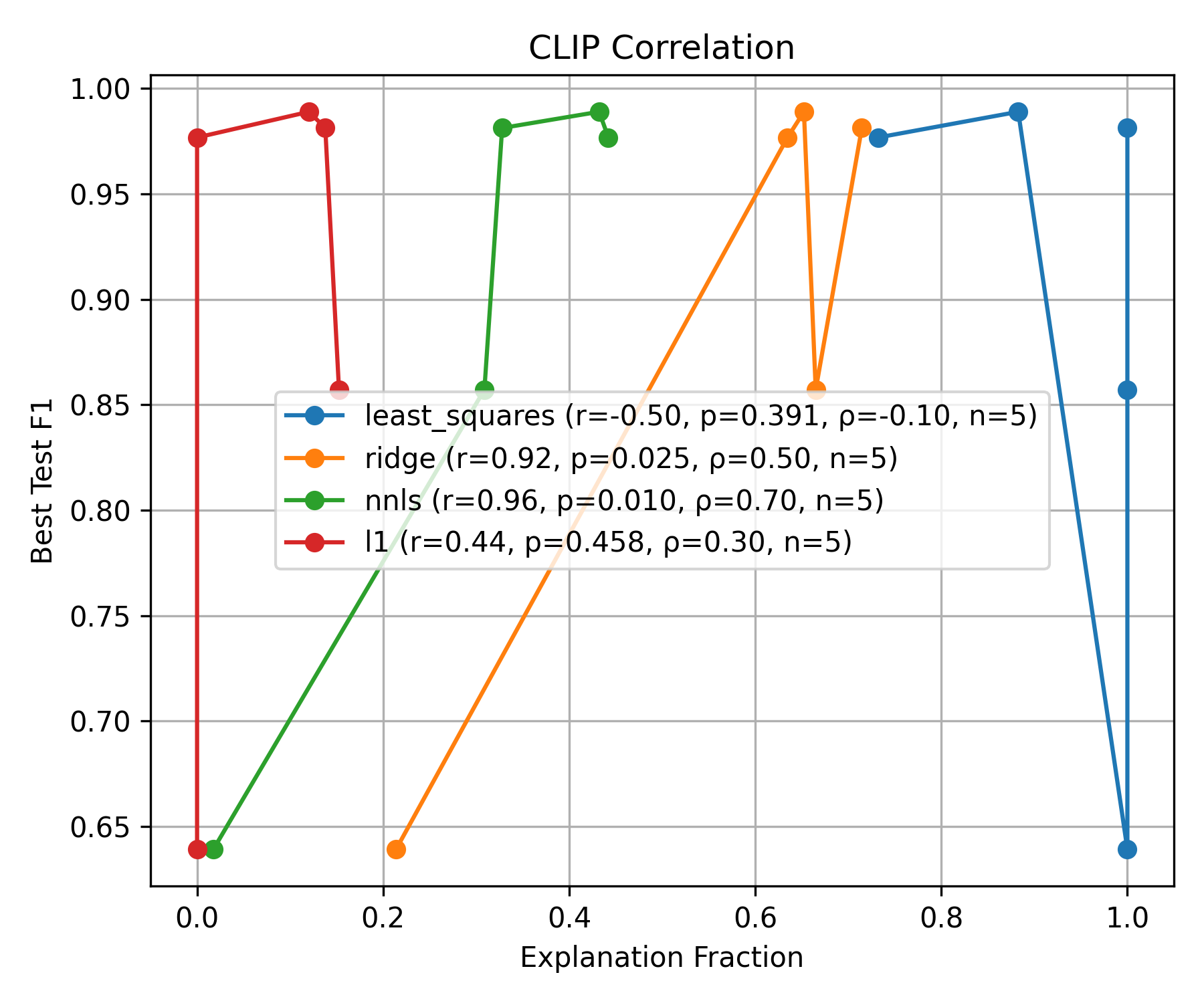}
        \text{(a) CLIP Embedding Space }
    \end{minipage}
    \hfill
    \begin{minipage}{0.48\textwidth}
        \centering
        \includegraphics[width=\linewidth]{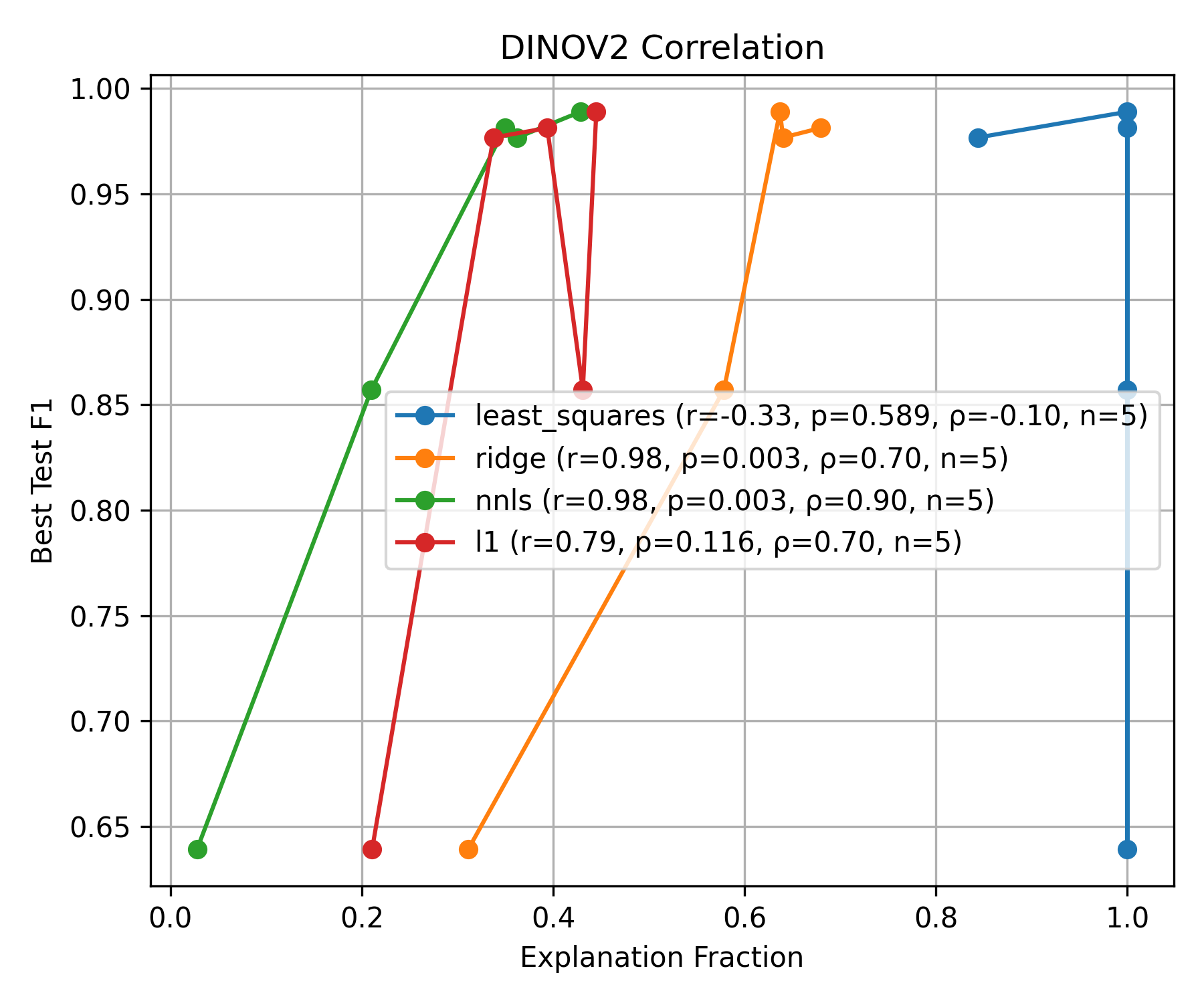}
        \text{(b) DINOv2 Embedding Space}
    \end{minipage}

    \vspace{0.5cm}

    \begin{minipage}{0.6\textwidth}
        \centering
        \includegraphics[width=\linewidth]{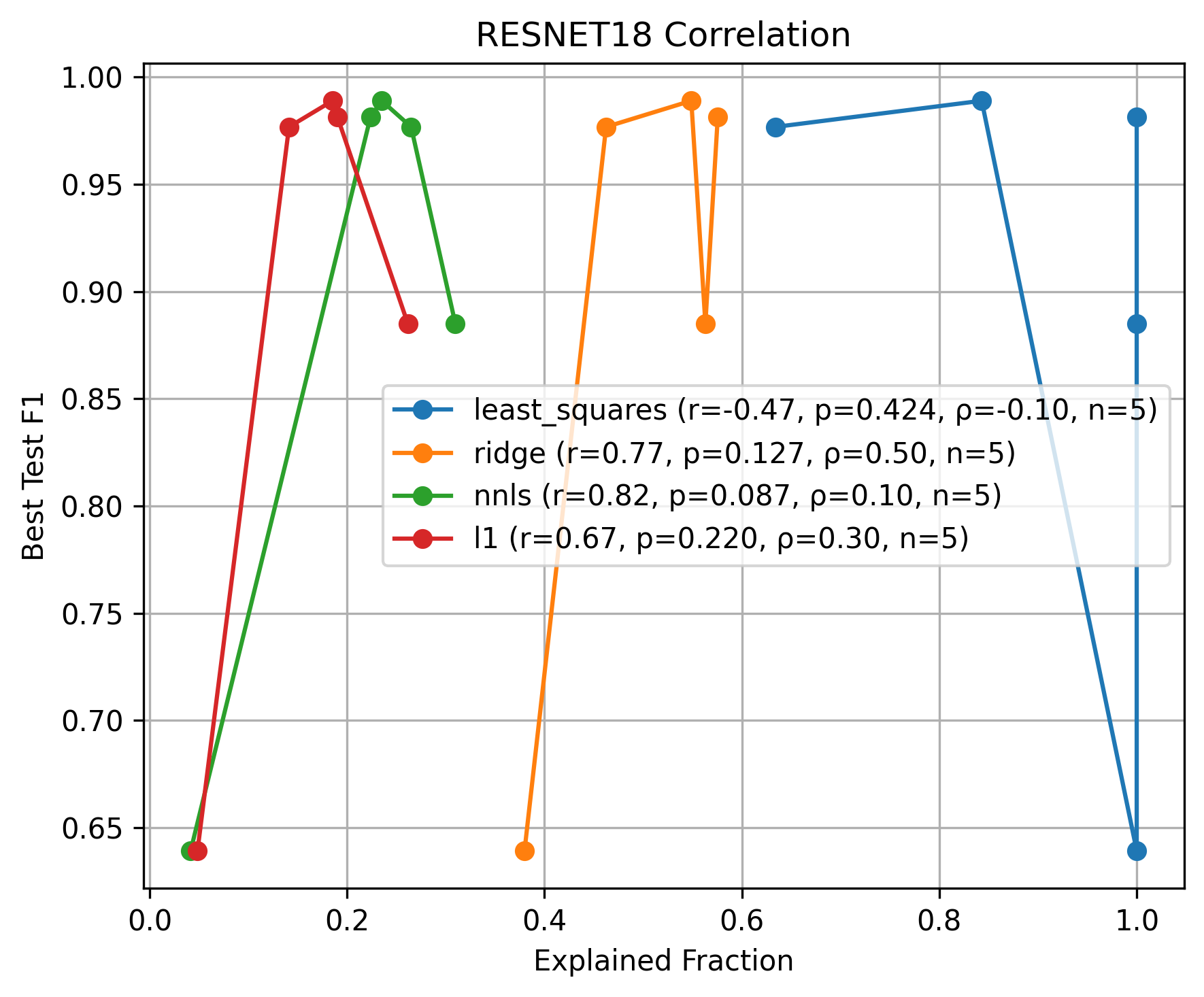}
        \text{(c) ResNet-18 Embedding Space}
    \end{minipage}

    \caption{
    \textbf{Correlation between Discriminative Span (DS) and downstream classification performance.} Higher DS values consistently map to improved downstream utility of synthetic data across (a) CLIP, (b) DINOv2, and (c) ResNet-18 embedding spaces. Note that regularized solvers successfully track downstream F1 scores, whereas unregularized Least Squares yields weak or negative correlations.
    }
    
    \label{fig:correlation_plots}
\end{figure*}

Figure~\ref{fig:correlation_plots} shows this relationship for CLIP embeddings, DINOv2 embeddings and ResNet-18 embeddings reports the results for ResNet-18. Each point corresponds to one dataset, allowing us to assess whether datasets that are easier to reconstruct also tend to yield stronger downstream performance.

The resulting trends provide a complementary view of the reconstruction estimator comparison. In particular, they allow us to distinguish an estimator that merely achieves low reconstruction error from one whose reconstruction quality is also aligned with the downstream utility of the learned representation. This correlation analysis is therefore used alongside the per-dataset reconstruction results rather than as a replacement for them.

\section{Validation of Vector Arithmetic in Visual Embeddings}
\label{app:vector_arithmetic}

Our method assumes that a visual transformation induces a reasonably
consistent directional change in embedding space. We validate this assumption
using a controlled transformation for which corresponding source and target
images are available.

\subsection{Experiment Design}

We construct a separate controlled toy dataset, distinct from the Toy
Watermark dataset used in our main experiments
(Appendix~\ref{app:toy_watermark}), from chest X-ray (CXR) images by
superimposing a red patch on the top-right quadrant of each image. We refer to this as the \textit{Toy Red-Patch CXR} dataset; it is used exclusively for validating the vector-arithmetic assumption in this section and is not used elsewhere in the paper. For each source-target pair, we extract embeddings using DINOv2, ResNet-18, and CLIP
and compute the corresponding difference vector
\[
\Delta_i =
f(x_i^{\mathrm{target}}) -
f(x_i^{\mathrm{source}}).
\]

We evaluate the directional consistency of these difference vectors using four
measures: (i) mean pairwise cosine similarity between difference vectors,
(ii) cosine similarity between each difference vector and their mean
direction, (iii) the variance explained by the leading principal components,
and (iv) the cosine similarity of the residual vectors after subtracting the
mean direction.

\subsection{Results}


\begin{table}[H]
\centering
\caption{Vector arithmetic alignment metrics on the Toy Red-Patch CXR dataset.}
\label{tab:vec_arith}
\begin{tabular}{lccc}
\toprule
\textbf{Metric} & \textbf{DINOv2} & \textbf{ResNet-18} & \textbf{CLIP} \\
\midrule
Mean cosine similarity              & 0.565   & 0.678   & 0.815   \\
Alignment with mean vector          & 0.753   & 0.824   & 0.903   \\
Variance explained by PC1           & 10.2\%  & 10.3\%  & 13.9\%  \\
Variance explained by first 3 PCs   & 25.2\%  & 23.1\%  & 29.8\%  \\
Residual cosine similarity          & $-$0.0029 & $-$0.0037 & $-$0.0032 \\
\bottomrule
\end{tabular}
\end{table}

The difference vectors exhibit positive directional alignment across all
three embedding spaces, with particularly strong alignment in CLIP. The
near-zero residual cosine similarity indicates that subtracting the dominant
mean direction substantially removes the shared directional component.
However, the leading principal component explains only $10$--$14\%$ of the
total variance, indicating that the transformation is not strictly
one-dimensional. Thus, the results support a more general view in which a
visual transformation induces a structured subspace containing a dominant
direction together with image-dependent variation. This provides empirical
support for representing synthetic transformations through the span of
paired embedding differences used by Discriminative Span.

\section{Geometry of the Transformation-Induced Subspace}
\label{app:geometry}

We look at whether simple global diversity metrics calculated from the difference matrix D are enough to explain why synthetic data works. Specifically, we analyze the effective rank, stable rank, and the condition number of $D$ across our datasets.

\subsection{Effective and Stable Rank.}

Let $\sigma_1,\ldots,\sigma_r$ denote the non-zero singular values of
$D$. The \textit{effective rank} provides a soft measure of the dimensionality
of the difference space by accounting for the distribution of singular values,
rather than simply counting non-zero singular values. Specifically, we define
the normalized singular-value distribution as
\[
p_i = \frac{\sigma_i}{\sum_{j=1}^{r}\sigma_j},
\]
and compute the effective rank as the exponential of its Shannon entropy:
\[
\mathrm{erank}(D)
=
\exp\left(
-\sum_{i=1}^{r} p_i \log p_i
\right).
\]
A larger effective rank indicates that the singular values are distributed
more uniformly across multiple directions, corresponding to a more
high-dimensional difference space.

The \textit{stable rank} is defined as
\[
\mathrm{srank}(D)
=
\frac{\|D\|_F^2}{\|D\|_2^2}
=
\frac{\sum_{i=1}^{r}\sigma_i^2}{\sigma_1^2},
\]
and measures the amount of total squared variation relative to the dominant
singular direction. Both effective rank and stable rank capture aspects of
global diversity in the transformation-induced variations, but neither
incorporates the downstream classification task.

\subsection{Conditioning.}
We additionally examine the condition number
\[
\kappa(D) = \frac{\sigma_{\max}(D)}{\sigma_{\min}(D)},
\]
together with the minimum singular value $\sigma_{\min}(D)$. Large condition
numbers and small minimum singular values indicate strong anisotropy and
near-linear dependencies among the difference vectors. These statistics are
particularly relevant to our setting because the reconstruction problem
operates in a high-dimensional embedding space and can therefore be
sensitive to poorly conditioned directions.

\subsection{Empirical Analysis of Difference-Space Geometry}

Tables~\ref{tab:rank_clip}--\ref{tab:rank_dino} report the effective rank,
stable rank, condition number, and minimum singular value of the difference
matrix across the three embedding spaces. The effective rank varies
substantially across datasets and embedding models, indicating that the
dimensionality of the transformation-induced variation is not uniform.
However, effective rank alone does not fully characterize the structure of
the difference space: difference matrices with comparable effective ranks can
exhibit substantially different stable ranks and conditioning. As reported in Section~\ref{sec:geometry_results}, the effective rank shows essentially no correlation with downstream F1 in the CLIP ($r\approx -0.01$) and DINOv2 ($r\approx 0.02$) embedding spaces, and a moderate negative correlation in the ResNet-18 embedding space ($r\approx -0.31$), underscoring that the raw dimensionality of the difference space, by itself, carries little information about downstream utility.

The singular-value structure provides a complementary view of this
variation. In the CLIP and DINOv2 embedding spaces, several difference
matrices exhibit large condition numbers together with very small minimum
singular values, indicating strong anisotropy and the presence of weak or
highly dependent directions. In contrast, the ResNet-18 embedding space
exhibits a different conditioning profile, with substantially smaller
condition numbers for several datasets.

These observations highlight that the dimensionality of the difference space
and the organization of its variation are distinct properties. A
transformation may induce substantial variation across the embedding space
without that variation necessarily being concentrated in directions useful
for the downstream task. Thus, rank and conditioning provide useful
task-independent geometric diagnostics, but they do not explicitly measure
task-relevant alignment. This motivates the use of Discriminative Span as a
complementary, task-aware measure of the transformation-induced subspace.

\begin{table}[H]
\centering
\caption{Rank and conditioning statistics of the difference matrix $D$ (CLIP embedding).}
\label{tab:rank_clip}
\begin{tabular}{lcccc}
\toprule
\textbf{Dataset} & \textbf{Eff.\ Rank} & \textbf{Stable Rank} & \textbf{Cond.\ Number} & \textbf{Min Singular} \\
\midrule
Pneumonia CXR   & 246.02 & 3.11 & $9.4 \times 10^{4}$ & 0 \\
Skin Lesion     & 191.6 & 2.79 & $3.1 \times 10^{2}$ & 0.19 \\
Toy Watermark   & 239.9 & 1.62 & $3.2 \times 10^{3}$ & 0.05 \\
Horses/Zebras   & 307.8 & 2.42 & $7.5 \times 10^{4}$ & 0 \\
Apples/Oranges  & 318.7 & 3.27 & $6.5 \times 10^{4}$ & 0 \\
\bottomrule
\end{tabular}
\end{table}

\begin{table}[H]
\centering
\caption{Rank and conditioning statistics of the difference matrix $D$ (ResNet-18 embedding).}
\label{tab:rank_resnet}
\begin{tabular}{lcccc}
\toprule
\textbf{Dataset} & \textbf{Eff.\ Rank} & \textbf{Stable Rank} & \textbf{Cond.\ Number} & \textbf{Min Singular} \\
\midrule
Pneumonia CXR   & 298.09 & 5.25 & $2.65 \times 10^{2}$ & 5.8e-1 \\
Skin Lesion     & 176.91 & 2.59 & $6.2 \times 10^{2}$ & 1.74e-1 \\
Toy Watermark   & 265.3 & 2.33 & $1.5 \times 10^{3}$ & 1.87e-1 \\
Horses/Zebras   & 320.51 & 2.38 & $1.7 \times 10^{2}$ & 2.19e+0 \\
Apples/Oranges  & 327.89 & 6.48 & $9.8 \times 10^{1}$ & 2.31e+0 \\
\bottomrule
\end{tabular}
\end{table}

\begin{table}[H]
\centering
\caption{Rank and conditioning statistics of the difference matrix $D$ (DINOv2 embedding).}
\label{tab:rank_dino}
\begin{tabular}{lcccc}
\toprule
\textbf{Dataset} & \textbf{Eff.\ Rank} & \textbf{Stable Rank} & \textbf{Cond.\ Number} & \textbf{Min Singular} \\
\midrule
Pneumonia CXR   & 232.34 & 3.66 & $1.8 \times 10^{7}$ & 0 \\
Skin Lesion     & 187.23 & 2.64 & $8.9 \times 10^{2}$ & 0.432 \\
Toy Watermark   & 249.89 & 4.27 & $5.1 \times 10^{7}$ & 0 \\
Horses/Zebras   & 263.45 & 2.00 & $5.8 \times 10^{7}$ & 0 \\
Apples/Oranges  & 267.68 & 2.95 & $6.2 \times 10^{7}$ & 0 \\
\bottomrule
\end{tabular}
\end{table}

\section{Per-Dataset Results}
\label{app:datasets_results}

The main paper reports the Pearson correlations between each synthetic-data
evaluation metric and downstream F1 score across datasets. Here, we provide
the underlying dataset-wise metric values used to compute these correlations.
These results provide the complete numerical breakdown for the three
representation spaces and allow the correlations reported in
Table~\ref{tab:correlation_btw_metrics_and_best_f1} to be reproduced directly.
We report Fr\'echet Distance, embedding similarity, linear-probe accuracy, and
Discriminative Span for each dataset and embedding space in Table ~\ref{tab:combined_metrics_results}. The effective-rank values used to compute the corresponding row of Table~\ref{tab:correlation_btw_metrics_and_best_f1} are reported separately in Appendix~\ref{app:geometry} (Tables~\ref{tab:rank_clip}--\ref{tab:rank_dino}), matched to the same five datasets and the Best Test F1 column below.


\begin{table*}[!ht]
\centering
\caption{Evaluation metrics across datasets and embedding spaces.}
\label{tab:combined_metrics_results}
\begin{tabular}{llccccc}
\toprule
\textbf{Foundation} & \textbf{Dataset} & \textbf{FD} & \textbf{Embedding Sim} & \textbf{Probe Acc} & \textbf{DS} & \textbf{Best Test F1} \\
\midrule
\multirow{5}{*}{clip} & Apples/Oranges & 22.6175 & 0.934 & 0.8350 & 0.665 & 0.885 \\
 & Horses/Zebra & 16.5864 & 0.9577 & 0.9127 & 0.715 & 0.9812 \\
 & Pneumonia CXR & 5.3787 & 0.986 & 0.4872 & 0.214 & 0.6392 \\
 & Skin Lesion & 20.0784 & 0.939 & 0.9929 & 0.635 & 0.9767 \\
 & Watermark Dataset & 8.2476 & 0.9966 & 0.9966 & 0.652 & 0.9889 \\
 \hline
\multirow{5}{*}{dinov2} & Apples/Oranges & 1343.1003 & 0.5941 & 0.6585 & 0.578 & 0.885 \\
 & Horses/Zebra & 481.1588 & 0.9314 & 0.9859 & 0.679 & 0.9812 \\
 & Pneumonia CXR & 402.2555 & 0.9542 & 0.1482 & 0.311 & 0.6392 \\
 & Skin Lesion & 831.4324 & 0.8684 & 0.9693 & 0.640 & 0.9767 \\
 & Watermark Dataset & 344.7316 & 0.9706 & 0.9813 & 0.637 & 0.9889 \\
\hline
\multirow{5}{*}{resnet18} & Apples/Oranges & 75.5687 & 0.9685 & 0.7775 & 0.563 & 0.885 \\
 & Horses/Zebra & 48.1439 & 0.9809 & 0.9705 & 0.576 & 0.9812 \\
 & Pneumonia CXR & 140.5901 & 0.9535 & 0.3748 & 0.380 & 0.6392 \\
 & Skin Lesion & 70.2398 & 0.9415 & 0.9433 & 0.462 & 0.9767 \\
 & Watermark Dataset & 36.7294 & 0.9892 & 0.9558 & 0.549 & 0.9889 \\
\bottomrule
\end{tabular}
\end{table*}


\begin{thebibliography}{20}
\bibitem{zhu2017unpaired}
Zhu, Jun-Yan, et al. "Unpaired image-to-image translation using cycle-consistent adversarial networks." Proceedings of the IEEE international conference on computer vision. 2017.
\bibitem{wolterink2017deep}
Wolterink, Jelmer M., et al. "Deep MR to CT synthesis using unpaired data." International workshop on simulation and synthesis in medical imaging. Cham: Springer International Publishing, 2017.
\bibitem{figueira2022survey}
A. Figueira and B. Vaz, "Survey on Synthetic Data Generation, Evaluation Methods and GANs," Mathematics, vol. 10, no. 15, p. 2733, 2022, doi: 10.3390/math10152733.
\bibitem{dankar2022multidimensional}
Dankar, FDa K., Mahmoud K. Ibrahim, and Leila Ismail. "A multi-dimensional evaluation of synthetic data generators." IEEE Access 10 (2022): 11147-11158.
\bibitem{wang2022dccyclegan}
J. Wang et al., "DC-cycleGAN: Bidirectional CT-to-MR Synthesis from Unpaired Data," arXiv preprint arXiv:2211.01293, 2022.
\bibitem{ibrahim2025generative}
Ibrahim, Mahmoud, et al. "Generative AI for synthetic data across multiple medical modalities: A systematic review of recent developments and challenges." Computers in biology and medicine 189 (2025): 109834.
\bibitem{koetzier2024synthetic}
Koetzier, Lennart R., et al. "Generating synthetic data for medical imaging." Radiology 312.3 (2024): e232471.
\bibitem{abdusalomov2023evaluating}
A. B. Abdusalomov et al., "Evaluating Synthetic Images Using Artificial Intelligence with the GAN Algorithm," Sensors, vol. 23, no. 7, p. 3440, 2023.
\bibitem{mumuni2024survey}
Alhassan, Mumuni, Fuseini Mumuni, and N. Gerrar. "A survey of synthetic data augmentation methods in computer vision." arXiv preprint (2024).
\bibitem{zamzmi2025scorecard}
Zamzmi, Ghada, et al. "Scorecard for synthetic medical data evaluation." Communications Engineering 4.1 (2025): 130.
\bibitem{sizikova2024synthetic}
Sizikova, Elena, et al. "Synthetic data in radiological imaging: current state and future outlook." BJR| Artificial Intelligence 1.1 (2024): ubae007.
\bibitem{radford2015unsupervised}
Radford, Alec, Luke Metz, and Soumith Chintala. "Unsupervised representation learning with deep convolutional generative adversarial networks." arXiv preprint arXiv:1511.06434 (2015).
\bibitem{lee2018diverse}
Lee, Hsin-Ying, et al. "Diverse image-to-image translation via disentangled representations." Proceedings of the European conference on computer vision (ECCV). 2018.
\bibitem{vecgan2022}
Dalva, Yusuf, Said Fahri Altındiş, and Aysegul Dundar. "Vecgan: Image-to-image translation with interpretable latent directions." European conference on computer vision. Cham: Springer Nature Switzerland, 2022.
\bibitem{slidergan2020}
Ververas, Evangelos, and Stefanos Zafeiriou. "Slidergan: Synthesizing expressive face images by sliding 3d blendshape parameters." International Journal of Computer Vision 128.10 (2020): 2629-2650.
\bibitem{chen2020simple}
Chen, Ting, et al. "A simple framework for contrastive learning of visual representations." International conference on machine learning. PMLR, 2020.

\bibitem{kermany2018} Kermany, Daniel S., et al. "Identifying medical diagnoses and treatable diseases by image-based deep learning." cell 172.5 (2018): 1122-1131.

\bibitem{ph2} Mendonça, Teresa, et al. "PH 2-A dermoscopic image database for research and benchmarking." 2013 35th annual international conference of the IEEE engineering in medicine and biology society (EMBC). IEEE, 2013.

\bibitem{isic} International Skin Imaging Collaboration (ISIC),
``ISIC Archive,'' available at \url{https://www.isic-archive.com/}.

\bibitem{jikadara_catsdogs}
B. Jikadara, ``Cats and Dogs Classification Dataset,''
Kaggle, available at
\url{https://www.kaggle.com/datasets/bhavikjikadara/dog-and-cat-classification-dataset}.

\end{thebibliography}
\end{document}